\documentclass[
twocolumn,
]{ceurart}

\usepackage{listings}
\usepackage{lipsum}
\usepackage{subcaption}
\usepackage{hyperref}
\usepackage{soul} 

\begin{document}

\copyrightyear{2023}
\copyrightclause{Copyright for this paper by its authors.
  Use permitted under Creative Commons License Attribution 4.0
  International (CC BY 4.0).}

\conference{SwissText'23: The 8th edition of the Swiss Text Analytics Conference -- Generative AI \& LLM, June 12--14, 2023, Neuchâtel, Switzerland}

\title{Large Language Model Prompt Chaining for\\Long Legal Document Classification}

\author[]{Dietrich Trautmann}[%
orcid=0000-0003-0858-6977,
email=Dietrich.Trautmann@tr.com,
]
\address[]{Thomson Reuters Labs, Zug, Canton of Zug, Switzerland}



\begin{abstract}
Prompting is used to guide or steer a language model in generating an appropriate response that is consistent with the desired outcome. 
Chaining is a strategy used to decompose complex tasks into smaller, manageable components. 
In this study, we utilize prompt chaining for extensive legal document classification tasks, which present difficulties due to their intricate domain-specific language and considerable length. 
Our approach begins with the creation of a concise summary of the original document, followed by a semantic search for related exemplar texts and their corresponding annotations from a training corpus. 
Finally, we prompt for a label - based on the task - to assign, by leveraging the in-context learning from the few-shot prompt. 
We demonstrate that through prompt chaining, we can not only enhance the performance over zero-shot, but also surpass the micro-F1 score achieved by larger models, such as ChatGPT zero-shot, using smaller models.
\end{abstract}


\maketitle

\section{Introduction}

The legal domain, with its often challenging tasks and complex long documents, is an important field of study for natural language processing (NLP) and machine learning \cite{dale2019law,zhong2020does}.
Long legal document text classification tasks can be challenging due to several factors, including large size, complex language and specific vocabulary, highly specialized content structure, imbalanced data (many common cases vs. a long-tail of peculiar ones), subjectivity (open to interpretation and debate), and the need for expensive, manual annotations from subject matter experts. 

The recent surge in the utilization of legal benchmarks has stimulated a proliferation of innovative solutions harnessing pre-trained language models \cite{chalkidis2022lexglue}.
Conventionally, these methodologies necessitate an intensive annotation process (though some utilize metadata annotations), followed by a costly fine-tuning process for the models \cite{chalkidis2022lexglue,chalkidis2021multieurlex}.

The advent of large-scale pre-training of large language models (LLMs) has presented an opportunity to leverage them directly through natural language prompting \cite{ouyang2022training}, circumventing the need for additional task-dependent fine-tuning.
Prompting involves providing a specific instruction, query, or question to an LLM to generate a specific output or response. 
The input, or prompt, steers the system towards producing a response meaningfully related to it.

\begin{figure}
  \includegraphics[width=\linewidth]{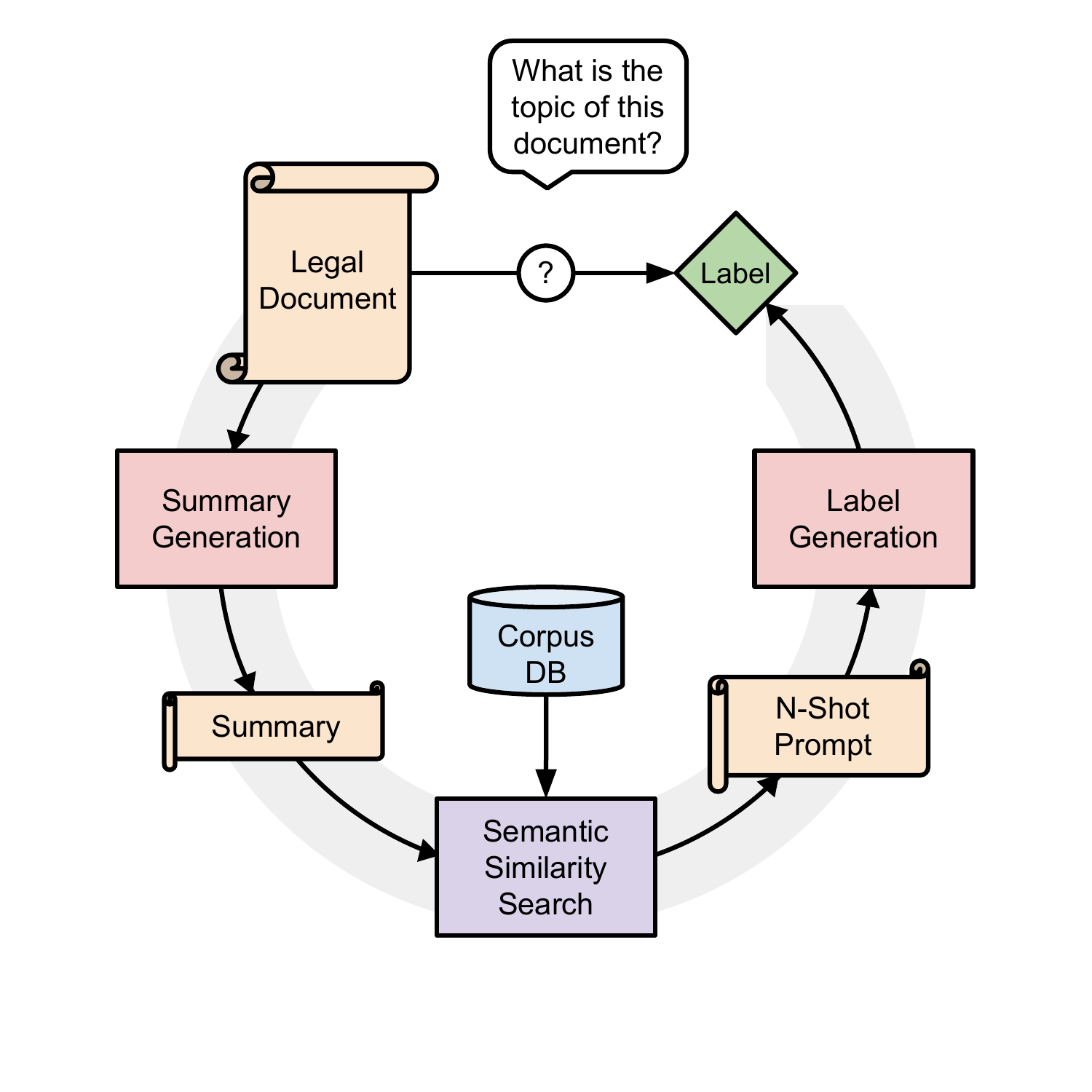}
  \caption{\emph{Prompt Chaining} for Legal Document Classification}
  \label{fig:prompt_chaining}
  \vspace*{-0.5cm}
\end{figure}

The technique of prompt chaining \cite{wu2022ai,wei2022chain} has shown promise in NLP, sequentially linking multiple prompts to guide the generation process (Fig. \ref{fig:prompt_chaining}). 
Through the utilization of consecutive prompts, the system can produce more contextually relevant responses for each step and more complex responses for the overall task.

Prompt chaining proves particularly advantageous in long legal document classification, improving task performance, efficiency, flexibility, and consistency via the inspection of individual steps in the chain \cite{wu2022promptchainer}. 
The technique enhances the interpretability of the overall classification and permits debugging of complex reasoning tasks upon failure \cite{khattab2022dsp}.

Overall, prompt chaining is a valuable tool in the classification of long legal documents, helping to improve the performance and efficiency of the classification process. 
Prompt chaining allows language models to build on their previous outputs and provide more nuanced classification results, and it can be customized to meet the specific needs of the legal document classification task.

Our contributions in this work are:
\begin{itemize}
    \item We show that we can successfully chain prompts for legal document classification tasks (visualized in Fig. \ref{fig:prompt_chaining}).
    \item We apply prompt chaining on one binary classification task and on one multi-class text classification task.
    \item We improve the results over zero-shot prompting with our chaining approach.
    \item Our prompt chaining approach even outperforms zero-shot ChatGPT prompting on the micro-f1 score.
\end{itemize}

\section{Related Work}

In terms of related literature we focus on the legal document classification work and on current prompting approaches, as well as the combination of the two fields.

\subsection{Legal Document Classification}

Documents, with their characteristic long textual data, often pose significant challenges for automated machine learning methods in processing and classification tasks \cite{wagh2021comparative}, \cite{park2022efficient}. 
These challenges become more pronounced in the legal domain due to additional complexities such as intricate grammar, nested sentences, domain-specific vocabulary, and extensive use of abbreviations \cite{garimella2022text}.

The \emph{LexGLEU} benchmark \cite{chalkidis2022lexglue} represents a comprehensive consolidation of recent datasets involving long legal documents, exclusively in the English language. 
It includes legal documents related to EU \& US Law, as well as contracts with tasks pertaining to multi-label and multi-class classification and multiple choice question-answering. 
In \cite{chalkidis2022exploration}, the authors evaluated a hierarchical approach for modeling long documents, and \cite{mamakas2022processing} investigated strategies to augment the context-window of transformers for domain-specific tasks, including the aforementioned \emph{LexGLEU} benchmark. 
While benchmarks in other languages do exist, such as \emph{MultiEURLEX} \cite{chalkidis2021multieurlex}, our work will also focus solely on the English language.

\subsection{Prompting}

Several noteworthy projects aim to consolidate, evaluate, and standardize prompting approaches across diverse tasks and domains. 
The two most substantial projects include OpenPrompt \cite{ding2022openprompt} and PromptSource \cite{bach2022promptsource}.

\paragraph{OpenPrompt} provides a user-friendly and research-friendly toolkit for prompt-learning in pre-trained language models (PLMs). 
The advent of prompt-learning in natural language processing sparked a need for a standardized implementation framework. 
OpenPrompt caters to this need by delivering a modular and extendable toolkit that accommodates various PLMs, task formats, and prompting modules in a unified paradigm.

\paragraph{PromptSource} is a toolkit designed to facilitate the development and sharing of natural language prompts for training and querying language models in NLP. 
It offers a templating language for creating data-linked prompts, a swift iterative development interface, and a community-driven set of guidelines for contributing new prompts. 
Currently, the platform offers over 2,000 prompts for approximately 170 datasets, promoting collaboration and efficient utilization of prompts for language model training and querying.

\paragraph{Prompt Chaining} as a concept was explored in \cite{wu2022promptchainer}, where prompts were chained using a visual program editor. 
Now, there are frameworks with prompt chaining at their core, such as LangChain \cite{chase2022langchain}, LLamaIndex \cite{liu2022llamaindex}, and MiniChain \cite{rush2023minichain}. 

\subsection{Legal Prompting}

Recent research has combined prompting and natural language tasks in the legal domain. 
In \cite{trautmann2022legal}, authors evaluated zero-shot prompting on the legal judgment prediction task using multilingual data from the European Court for Human Rights (ECHR) and the Federal Supreme Court of Switzerland (FSCS). 
Meanwhile, \cite{yu2022legal} appraised \emph{GPT-3}'s zero- and few-shot capabilities for legal reasoning tasks on the COLLIE entailment task (using English translations of the Japanese bar exam).

The GPT-3.5 model was evaluated on the US Bar Exam \cite{bommarito2022gpt}, and GPT-4 \cite{katz2023gpt} has demonstrated proficiency in passing multiple tests through zero-shot prompting.

\section{Data}

The datasets utilized in our study are sourced from two widely recognized benchmarks comprising lengthy documents in the legal domain: the European Court of Human Rights (ECHR) and the Supreme Court of the United States (SCOTUS). 
These datasets form part of the \emph{LexGLUE} benchmark \cite{chalkidis2022lexglue}.

\subsection{ECHR}

The ECHR dataset comprises approximately $11,000$ cases sourced from the European Court of Human Rights public database. 
This dataset is split into training, development, and test sets. 
Each case includes factual paragraphs, along with the corresponding ECHR articles that were violated or alleged to be violated. 
The original task involves to predict the violated human rights articles for a case's facts. 
However, for the purpose of our study, we have simplified this task to a binary classification problem. 
We distinguish cases based on whether there was a violation of any human rights articles, irrespective of which specific articles were violated.

\subsection{SCOTUS}

The SCOTUS dataset provides insight into the highest federal court in the USA, which handles complex or controversial cases unresolved by lower courts. 
This dataset combines information from SCOTUS opinions and the Supreme Court DataBase (SCDB). 
The SCDB offers metadata for all cases spanning from 1946 to 2020. 
Utilizing the SCDB, the dataset classifies court opinions into 14 issue areas (refer to App. \ref{app:scotus_issue_areas}). 
The dataset is divided chronologically into training (\#5000 samples, 1946–1982), development (\#1400 samples, 1982–1991), and test (\#1400 samples, 1991–2016) sets, each covering a distinct time period.

\section{Models}

Our experimental design incorporates both general-purpose text generation models and task-specific summarization models.

\subsection{Generation Models}

We used two different 20 billion parameter LLMs in our text generation steps.
Both of the models have a context window of up to 2048 tokens.

\subsubsection{GPT-NeoX}
The \emph{GPT-NeoX} model \cite{black2022gpt} is an autoregressive language model, specifically a decoder-only model\footnote{\url{https://hf.co/EleutherAI/gpt-neox-20b}}, trained on the Pile dataset \cite{gao2020pile}. 
This models' weights are openly available under a permissive license (Apache 2.0 \footnote{\label{apache_license}\url{https://www.apache.org/licenses/LICENSE-2.0.html}}).

\subsubsection{Flan-UL2}
\emph{Flan-UL2} \cite{tay2022ul2} is an encoder-decoder model\footnote{\url{https://hf.co/google/flan-ul2}} based on the T5 architecture, trained with the mixture-of-denoisers objectives (diverse span corruption and prefix language modeling tasks). 
This model was further instruction fine-tuned using the Flan prompting collection \cite{longpre2023flan}.
The collection contains instructions for a diverse set of tasks (e.g., to summarize a text; to classify based on a list of options).
The model is publicly available with an open source license (Apache 2.0).

\subsection{Summarization Models}

We also used task-specific summarization models for the creation of the legal summaries.
In our experiments we found that -- due to the lack of ground-truth summaries for our long legal documents -- the task-specific summarization models created more coherent summaries compared to results from prompting the general generation models.

\subsubsection{BRIO}

The BRIO\footnote{\url{https://hf.co/Yale-LILY/brio-cnndm-uncased}} model \cite{liu2022brio} is an abstractive summarization model that achieves state-of-the-art result on the \emph{CNN/DailyMail} and \emph{XSum} datasets.
It uses BART \cite{lewis2020bart} as its base model and has a context window of 1024 tokens.

\subsubsection{PRIMERA}

The PRIMERA\footnote{\url{https://hf.co/allenai/primera-multi_lexsum-source-short}} model \cite{shen2022multi} is an abstractive summarization model that was trained on the \emph{Multi-LexSum} dataset at different granularities.
We used the model that was trained on the granularity from the full source document to create a short summary and it has a context window of 1024 tokens.
The other options are -- besides different model architectures -- long and tiny summaries.

\subsection{Semantic Similarity Search}

We use semantic similarity search for the few-shot prompt building were we retrieve semantic similar summaries from the training set to a target summary (either from the development or test set).
For this purpose, our summaries were encoded using the\emph{sentence-transformers} library \cite{reimers-2019-sentence-bert} and the \emph{custom-legalbert}\footnote{\url{https://hf.co/zlucia/custom-legalbert}} model \cite{zheng2021does}.
Furthermore, we used the \emph{annoy}\footnote{\url{https://github.com/spotify/annoy}} library for the approximate nearest neighbor search (semantic similarity search).

\section{Prompt Chaining}

Prompt Chaining is a methodology employed to decompose complex tasks into smaller, manageable sub-tasks. 
A prompt chain typically comprises several prompts - either task-specific or general-purpose, each serving a single purpose.
The output of one prompt feeds into the next as an input. 
In our approach, we utilize pre-defined steps; however, it's worth noting that this methodology could be further optimized by incorporating a selection process for the next step or by introducing stopping criteria in the output generation, as exemplified in \cite{khattab2022dsp} and \cite{schick2023toolformer}. 
The steps for our prompt chaining process, specifically utilized for the classification of long legal documents, are depicted in Fig. \ref{fig:prompt_chaining}. 
In the following sections, we will delve into each of the primary steps, namely summarization, few-shot prompt building, and final label generation.

\subsection{Summary Generation}

The inaugural step in our prompt chaining approach is the generation of a succinct summary of the legal case text. 
As a majority of legal documents are lengthy, often exceeding the large context window of $2048$ tokens provided by contemporary language models, we advocate the creation of summaries for chunks of the whole document. 
These chunks are crafted based on the model's context window. 
Sentences are sequentially stacked until the context limit of the respective models is reached ($1024$ or $2048$ tokens).
Post the initial parsing of the full document, the summary generation process for chunks is iteratively continued until the desired summary length, in our case up to $128$ tokens, is achieved. 
These summaries typically consist of a few sentences (up to 5) derived from our data.

Initial experimentation with direct prompting approaches on large language models resulted in variable outcomes. 
The templates used for prompting included:

\begin{itemize}
    \item {{INPUT TEXT}} \emph{In summary, }
    \item {{INPUT TEXT}} \emph{TLDR: }
\end{itemize}

Since we do not possess ground-truth summaries for the documents, our assessments relied on manual inspection of a subset of the summaries generated (from the training set). 
The inspection indicated that the summaries were relatively generic and often omitted the core legal issues of interest.

Consequently, our investigation steered towards task-specific summarization models. 
Notably, the \emph{BRIO} model, being pre-trained and fine-tuned on news articles, generated more generic summaries. 
In contrast, the \emph{PRIMERA} model, fine-tuned specifically on legal documents, generated summaries where the core legal context was mostly preserved. 
This iterative summarization was uniformly applied across all documents using the same parameters.

\subsection{Semantic Similarity Search}

The objective at this stage was to construct few-shot prompts for the subsequent label generation step. 
Thus, we embedded all our summaries with the models discussed in section 4.3 and calculated the semantically closest neighbors (up to eight) of each summary in the development and test sets, as compared to the training set. 
These summaries from the training set, along with their true labels and the current target sample (excluding its true label from the development and test sets), served as the few-shot prompts. 
This approach leverages the in-context learning abilities of large language models.

As illustrated by \cite{yu2022generate}, an alternative approach could involve prompting a large language model to generate contextual texts based on an input, instead of retrieving from a corpus database as we have done. 
These generated samples could then be incorporated as in-context samples in the following step. 
This option can also be potentially implemented via LLMs prompting in our prompting pipeline. 
The evaluation of this approach's feasibility is left for future work.

\subsection{Label Generation}

The final step in our prompt chain involves label generation. 
Here, we queried the LLMs with the few-shot prompts previously constructed, in conjunction with an instruction for the corresponding task and a list of potential labels to choose from. 
For the ECHR experiments, the binary options provided were \emph{YES} or \emph{NO}, while for the SCOTUS experiments, up to 13 issue area labels were presented for model prediction. 
This prompt construction enabled the models to yield the desired label in all our experiments. 
Greedy decoding was used in this generation step for all results.

Another strategy employed involved sampling from the model output multiple times, a technique known as self-consistency via output sampling \cite{wang2022self}. 
It has been demonstrated that querying multiple times enhances the probability of generating the true label. 
At the conclusion of each such sampling (up to 10 times), the majority count of the generated labels was taken as the final prediction.

\section{Experiments}

Our experiments pursued two objectives. 
Firstly, we aimed to enhance the zero-shot outcomes from prior work on the binary classification task on the ECHR dataset, leveraging prompting techniques without any parameter adjustments to the models. 
Following the successful demonstration of the efficacy of the few-shot approach, we expanded our focus. 
The second experiment involved extending the process to the 13 labels in the SCOTUS corpus, a task significantly more challenging than the prior binary classification.

In addition to this, we compared our results to the zero-shot ChatGPT results reported in \cite{chalkidis2023chatgpt}, which covered a subset of the overall samples in the SCOTUS data. 
We selected the corresponding samples and provided our results for comparison.

Over multiple iterations, we developed the few-shot prompts on a randomly selected portion (n = 40) of the development sets for both the \emph{ECHR} and the \emph{SCOTUS} datasets. 
The final eight-shot prompt incorporated the eight semantically closest summaries from the training set to the corresponding target set, as well as the respective gold label from the training set. 
Notably for the \emph{SCOTUS} corpus, we limited the available issue area labels for the model, based on the labels of the included eight samples. 
Once we identified the most effective composition of the few-shot prompt on the random sample, we applied it to the full development and test sets, and reported the results.

Our computational requirements were limited to CPU resources, as we only performed inference calls on the generative models. 
Our work did not involve the use of any GPUs. 
Detailed computational information is available in the appendix (see App. \ref{app:compute}).

\section{Results}

In this section, we discuss our results on both benchmarks. 
We have included the confusion matrices for the development sets (see Fig. \ref{fig:cm_ecthr_dev} for ECHR and Fig. \ref{fig:cm_scotus_dev} for SCOTUS), the labelwise F1-scores (see Tab. \ref{tab:ecthr-labelwise-results} and Tab. \ref{tab:scotus-labelwise-results}), and the overall results (see Tab. 2 and Tab. \ref{tab:results_scotus}).

\subsection{ECHR Results}

\begin{figure}
  \centering
  \includegraphics[width=0.38\linewidth]{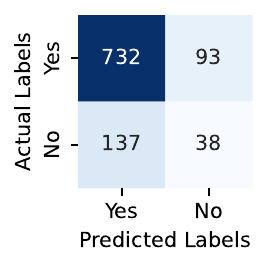}
  \caption{Confusion matrix for the dev. set of \emph{ECHR}.}
  \label{fig:cm_ecthr_dev}
\end{figure}

With respect to the ECHR results, we managed to improve upon the zero-shot results from previous work. 
However, the few-shot context still did not suffice to reach the performance of supervised trained models. 
It's important to remember that full supervised fine-tuning involves hours of update runs with thousands of annotated samples, while our experiments only included inference calls. 
The confusion matrix (Fig. \ref{fig:cm_ecthr_dev}) also demonstrates some misclassification along the off-diagonal axis, although the few-shot prompting did capture more of the minority (\emph{NO}) class.

\begin{table}[!h]
\begin{center}
    \begin{tabular}{r|l|r|c}
        \# & Label & \# Samples & F1 \\
        \hline
         1 & Yes & 825 & 0.864 \\
         2 & No  & 175 & 0.248 \\
    \end{tabular}
    \caption{Labelwise F1-Scores for the development set of \emph{ECHR}.}
\label{tab:ecthr-labelwise-results}
\end{center}
\end{table}

\newpage
\subsection{SCOTUS Results}

The results for the multi-class text classification task for \emph{SCOTUS} are presented in Tab. 2. 
Alongside our results on the development and test sets, we also included external (ext.) results from \cite{chalkidis2022lexglue} and \cite{chalkidis2023chatgpt}.

While we achieved satisfactory performance on the development set, we observed a substantial drop in performance on the test set. 
The test set performance in terms of macro-f1 score was below the zero-shot ChatGPT results. 
However, our prompt chaining approach was more effective in retrieving the higher frequency classes, as reflected in the better micro-f1 score.

\begin{figure}
  \includegraphics[width=\linewidth]{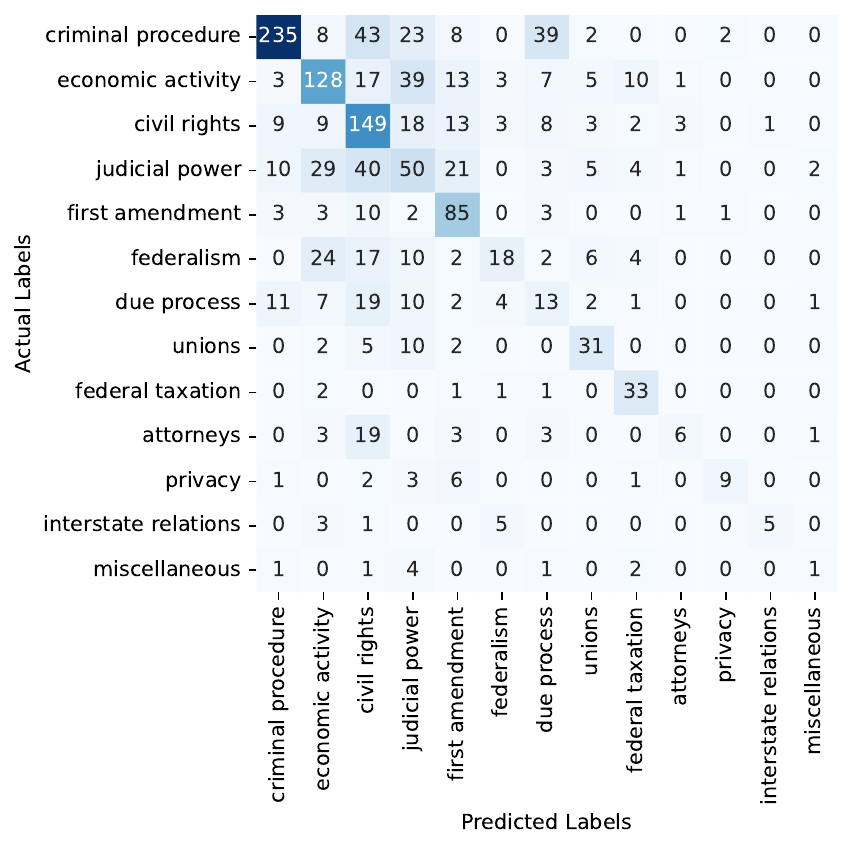}
  \caption{Confusion matrix for the dev. set of \emph{SCOTUS}.}
  \label{fig:cm_scotus_dev}
\end{figure}

This trend is also evident in the label-wise scores (Tab. \ref{tab:scotus-labelwise-results}), where the higher frequency classes received better scores than the minority classes. 
The confusion matrix (Fig. \ref{fig:cm_scotus_dev}) for this experiment showed that particularly many issue areas were predicted as \emph{civil rights}, while also the \emph{criminal procedure}, \emph{judicial power} and \emph{federalism} were misclassified as others.

\begin{table*}[!h]
    \centering
    \begin{tabular}{c|l|c|c|c||c|c|c}
        \toprule
        & \multicolumn{1}{c|}{Model} & \multicolumn{1}{c|}{Precision} & \multicolumn{1}{c|}{Recall} & \multicolumn{1}{c||}{macro-F1} & \multicolumn{1}{c|}{micro-F1} & \multicolumn{1}{c|}{weighted-F1} & \multicolumn{1}{c}{Accuracy} \\
        \midrule
        \parbox[t]{2mm}{\multirow{5}{*}{\rotatebox[origin=c]{90}{dev. set}}} &
          minority class         & .088 & .500 & .149 & .175 & .052 & .175 \\
        & random class           & .506 & .510 & .451 & .514 & .572 & .514 \\
        & majority class         & .412 & .500 & .452 & \textbf{.825} & \textbf{.746} & \textbf{.825} \\
        & GPT-NeoX (0-shot, (F)) & .527 & .536 & .526 & .709 & .731 & .709 \\
        & GPT-NeoX (8-shot, (S)) & \textbf{.566} & \textbf{.552 }& \textbf{.556} & .770 & .756 & .770 \\
        \midrule
        \parbox[t]{2mm}{\multirow{5}{*}{\rotatebox[origin=c]{90}{test set}}} &
          minority class         & .077 & .500 & .133 & .153 & .041 & .153 \\
        & random class           & .479 & .460 & .410 & .484 & .555 & .484 \\
        & majority class         & .423 & .500 & .459 & \textbf{.847} & \textbf{.777} & \textbf{.847} \\
        & GPT-NeoX (0-shot, (F)) & .522 & .530 & .521 & .707 & .728 & .707 \\
        & GPT-NeoX (8-shot, (S)) & \textbf{.525} & \textbf{.537} & \textbf{.527} & .779 & .768 & .779 \\
        \bottomrule
    \end{tabular}
    \caption{
        The results for the ECtHR development and test sets. 
        Besides the macro-averaged F1-score, precision and recall, we report also the micro-averaged and weighted-F1 and the accuracy scores. 
        The (F) stands for the full document that was used in the input to the model, while the (S) stands for the summaries (concatenated in the few-shot prompts) as the input to the model.
    }
  \label{tab:results_echr}
\end{table*}
\begin{table*}[!h]
    \centering
    
    \begin{tabular}{c|l|c|c|c||c|c|c}
        \toprule
        & \multicolumn{1}{c|}{Model} & \multicolumn{1}{c|}{Precision} & \multicolumn{1}{c|}{Recall} & \multicolumn{1}{c||}{macro-F1} & \multicolumn{1}{c|}{micro-F1} & \multicolumn{1}{c|}{weighted-F1} & \multicolumn{1}{c}{Accuracy} \\
        \midrule
        \parbox[t]{2mm}{\multirow{3}{*}{\rotatebox[origin=c]{90}{dev. set}}} &
          majority class         & .020 & .077 & .031 & .257 & .105 & .257 \\
        & random class           & .070 & .064 & .057 & .071 & .084 & .071 \\
        & FLAN-UL2 (8-shot, (S)) & \textbf{.529} & \textbf{.455} & \textbf{.461} & \textbf{.545} & \textbf{.543} & \textbf{.545} \\
        \midrule
        \parbox[t]{2mm}{\multirow{4}{*}{\rotatebox[origin=c]{90}{test set}}} &
          majority class         & .020 & .077 & .032 & .266 & .112 & .266 \\
        & random class           & .079 & .074 & .060 & .077 & .095 & .077 \\
        & FLAN-UL2 (8-shot, (S))            & .427 & .373 & .359 & \textbf{.486} & \textbf{.483} & \textbf{.486} \\
        & FLAN-UL2 (8-shot, (S))\textdagger & \textbf{.435} & \textbf{.388} & \textbf{.371} & .484 & .480 & .484 \\
        \midrule
        \parbox[t]{2mm}{\multirow{2}{*}{\rotatebox[origin=c]{90}{ext.}}} &
          ChatGPT (0-shot, (F))  & -    & -    & .420 & .438  & - & - \\
        & supervised (full)      & -    & -    & \textbf{.695} & \textbf{.782} & - & - \\
        \bottomrule
    \end{tabular}
    \caption{
        The results for the SCOTUS development and test sets. 
        Besides the macro-averaged F1-score, precision and recall, we report also the micro-averaged and weighted-F1 and the accuracy scores. 
        The (F) stands for the full document that was used in the input to the model, while the (S) stands for the summaries (concatenated in the few-shot prompts) as the input to the model.
        \textdagger We calculated the scores based on the same reduced set of documents (1k) as the ChatGPT work.
        The ext. rows are external results copied from the corresponding papers \cite{chalkidis2022lexglue,chalkidis2023chatgpt}.
    }
  \label{tab:results_scotus}
\end{table*}


\begin{table}[!h]
\begin{center}
    \begin{tabular}{r|l|r|c}
        \toprule
        \# & Label & \# Samples & F1 \\
        \hline
         1 & Criminal Procedure   &  360 & 0.742 \\
         2 & Federal Taxation     &  226 & 0.695 \\
         3 & First Amendment      &  218 & 0.644 \\
         4 & Unions               &  165 & 0.590 \\
         5 & Economic Activity    &  108 & 0.577 \\
         6 & Civil Rights         &   83 & 0.551 \\
         7 & Privacy              &   70 & 0.529 \\
         8 & Interstate Relations &   51 & 0.500 \\
         9 & Federalism           &   38 & 0.308 \\
        10 & Judicial Power       &   35 & 0.299 \\
        11 & Attorneys            &   22 & 0.255 \\
        12 & Due Process          &   14 & 0.173 \\
        13 & Miscellaneous        &   10 & 0.133 \\
        \bottomrule
    \end{tabular}
    \caption{Labelwise F1-Scores for the development set of \emph{SCOTUS}.}
\label{tab:scotus-labelwise-results}
\end{center}
\end{table}

\section{Conclusion}

Our experiments successfully demonstrated that the implementation of few-shot prompts can lead to improvements upon the zero-shot results. 
We also showed that it is feasible to predict frequent labels with appreciable F1 scores using this approach.

The strategy of prompting, and as we have demonstrated, the concept of prompt chaining, represent promising avenues for future exploration. 
These techniques are particularly advantageous as they circumvent the need for costly data annotation and the development of custom models.

Last but not least, established prompting pipelines can be adapted for use with different (updated) models and, as shown in \cite{katz2023gpt}, they offer across-the-board enhancements for a diverse range of tasks, free of cost. 
Looking ahead, our future work aims to experiment with even larger models on additional legal benchmarks.

\newpage
\bibliography{mybib}

\appendix

\section{Compute Requirements}
\label{app:compute}
We used the following Amazon EC2 M5 instance:

\vspace{1em}

\begin{tabular}{c|c|c} 
  \toprule
  \textbf{Standard Instance} & \textbf{vCPU} & \textbf{Memory} \\ 
  \midrule
  ml.m5d.24xlarge & 96 & 384 GiB \\ 
  \bottomrule
\end{tabular}

\vspace{1em}

\noindent We haven't used any GPUs in our experiments.

\section{SCOTUS issue areas}
\label{app:scotus_issue_areas}

\begin{itemize}
    \item Criminal Procedure
    \item Civil Rights
    \item First Amendment
    \item Due Process
    \item Privacy
    \item Attorneys
    \item Unions
    \item Economic Activity
    \item Judicial Power
    \item Federalism
    \item Interstate Relations
    \item Federal Taxation
    \item Miscellaneous
    \item \st{Private Action} was not available in the data
\end{itemize}

\end{document}